\documentclass[11pt]{article}
\usepackage[colorlinks,citecolor=gray,linkcolor=black,bookmarks=false]{hyperref} 

\usepackage[final]{acl}

\usepackage{times}
\usepackage{latexsym}

\usepackage[T1]{fontenc}

\usepackage[utf8]{inputenc}

\usepackage{microtype}

\usepackage[ruled,vlined]{algorithm2e}
\usepackage{tabularx}
\usepackage{makecell}
\usepackage{algpseudocode}
\usepackage{booktabs}
\usepackage{adjustbox}
\usepackage{hhline}
\usepackage{xspace}
\usepackage{xcolor}         
\usepackage{url}
\usepackage{graphicx}
\usepackage{wrapfig}
\usepackage{trajan}
\usepackage{capt-of}
\usepackage{multicol}
\usepackage{multirow}
\usepackage{inconsolata}
\usepackage{listings}
\usepackage{todonotes}
\usepackage[cachedir=.,frozencache]{minted}

\usepackage{graphicx}
\usepackage{cleveref}
\usepackage{verbatim}
\usepackage[labelformat=simple]{subcaption}

\usepackage{wasysym}
\usepackage{hyperref}

\author{Ruohong Zhang$^{\spadesuit}$\thanks{\quad Correspondence: \texttt{ruohongz@andrew.cmu.edu}. Work in progress.}  \quad Luyu Gao$^{\spadesuit}$ \quad Chen Zheng$^{\eighthnote}$ \quad Zhen Fan$^{\spadesuit}$ \\ {\bf Guokun Lai}$^{\eighthnote}$ \quad
{\bf Zheng Zhang}$^{\clubsuit}$ \quad {\bf Fangzhou Ai}$^{\diamondsuit}$ \quad
{\bf Yiming Yang}$^{\spadesuit}$ \quad {\bf Hongxia Yang}$^{\eighthnote}$ \\
   $^{\spadesuit}$CMU, $^{\clubsuit}$Emory University, $^{\diamondsuit}$UC San Diego, $^{\eighthnote}$TikTok
}

\newcommand{\llminer}{\textsc{LLMiner}\xspace}

\definecolor{ruohong}{RGB}{101, 66, 243}
\definecolor{yiming}{RGB}{255, 0, 0}

\definecolor{CBF1}{RGB}{255,99,132}  %
\definecolor{CBF2}{RGB}{54,162,235}  %
\definecolor{CBF3}{RGB}{255,206,86}  %
\definecolor{CBF4}{RGB}{75,192,192}  %
\definecolor{CBF5}{RGB}{153,102,255} %
\definecolor{CBF1b}{RGB}{205,89,112}  %
\definecolor{CBF2b}{RGB}{44,142,215}  %
\definecolor{CBF5b}{RGB}{133,92,225}  %

\usepackage{soul}

\newcommand{\squishlist}{
  \begin{list}{$\bullet$}
    { \setlength{\itemsep}{0pt}      \setlength{\parsep}{3pt}
      \setlength{\topsep}{3pt}       \setlength{\partopsep}{0pt}
      \setlength{\leftmargin}{1.5em} \setlength{\labelwidth}{1em}
      \setlength{\labelsep}{0.5em} } }
\newcommand{\reallysquishlist}{
  \begin{list}{$\bullet$}
    { \setlength{\itemsep}{0pt}    \setlength{\parsep}{0pt}
      \setlength{\topsep}{0pt}     \setlength{\partopsep}{0pt}
      \setlength{\leftmargin}{0.2em} \setlength{\labelwidth}{0.2em}
      \setlength{\labelsep}{0.2em} } }

 \newcommand{\squishend}{
     \end{list} 
 }

\renewcommand{\cite}{\citep}

\definecolor{lightgray}{gray}{0.9}

\definecolor{Box1Color}{RGB}{227, 236, 246}
\definecolor{Box2Color}{RGB}{248, 220, 225}
\definecolor{Box3Color}{RGB}{255, 238, 224}

\definecolor{cbBlue}{RGB}{0, 114, 178}
\definecolor{cbOrange}{RGB}{240, 228, 66}
\definecolor{cbGreen}{RGB}{0, 158, 115}
\definecolor{cbRed}{RGB}{213, 94, 0}
\definecolor{cbPurple}{RGB}{204, 121, 167}
\definecolor{cbSkyBlue}{RGB}{86, 180, 233}
\definecolor{cbGray}{RGB}{128, 128, 128}

\newcommand{\num}{600\xspace}

\begin{document}

\title{A Self-enhancement Approach for Domain-specific Chatbot Training via Knowledge Mining and Digest}
%
%
%
\maketitle              
\begin{abstract}

Large Language Models (LLMs), despite their great power in language generation, often encounter challenges when dealing with intricate and knowledge-demanding queries in specific domains. 
This paper introduces a novel approach to enhance LLMs by effectively extracting the relevant knowledge from domain-specific textual sources, and the adaptive training of a chatbot with domain-specific inquiries. 
Our two-step approach starts from training a knowledge miner, namely \llminer, which autonomously extracts Question-Answer pairs from relevant documents through a chain-of-thought reasoning process. 
Subsequently, we blend the mined QA pairs with a conversational dataset to fine-tune the LLM as a chatbot, thereby enriching its domain-specific expertise and conversational capabilities.  
We also developed a new evaluation benchmark which comprises four domain-specific text corpora and associated human-crafted QA pairs for testing. Our model shows remarkable performance improvement over generally aligned LLM and surpasses domain-adapted models directly fine-tuned on domain corpus. 
In particular, \llminer achieves this with minimal human intervention, requiring only \num seed instances, thereby providing a pathway towards self-improvement of LLMs through model-synthesized training data.
\end{abstract}

\section{Introduction}
\begin{figure}[t!]
    \centering
    \includegraphics[width=0.9\linewidth]{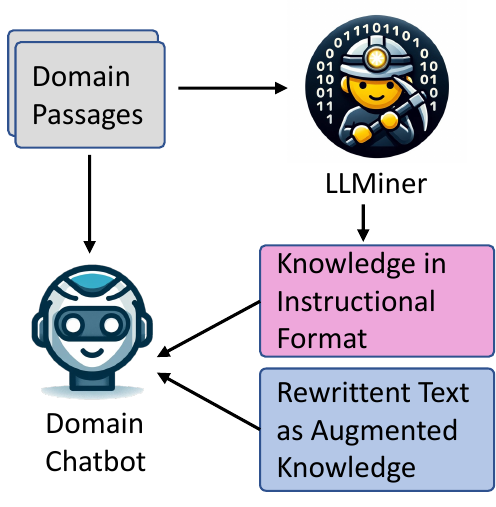}
    \caption{Illustration of the \llminer approach for enhancing Large Language Models (LLMs) as domain-specific chatbots.}
    \label{fig:illustrative}
\end{figure}

Large Language Models (LLMs) have significantly advanced the fields of natural language understanding and generation. Despite these advancements, training LLMs as dynamic AI agents capable of answering complex, domain-specific, and knowledge-intensive questions remains a significant challenge~\cite{abbasian2023conversational, ge2023openagi}. Current methods to integrate knowledge into LLMs primarily involve either modifying model weights with manually curated knowledge~\cite{hase2021language, meng2022generating, mitchell2022memory, dai2021knowledge} or continuous pre-training with a language modeling objective on domain-specific texts \citep{jang2021towards, jin2021lifelong, qin2022elle}. However, these methods often fall short in addressing complex queries that extend beyond simple fact retrieval \citep{zhong2023mquake}, and they tend to rely on the mere memorization~\cite{zhu2023physics, wang2021can} of factual content, which proves insufficient for delivering insightful responses to specialized questions.

This paper introduces a novel approach for empowering LLMs with domain-specific expertise through an autonomous knowledge mining process (\cref{fig:illustrative}). We propose \llminer, an LLM agent that autonomously extracts knowledge directly from domain-specific texts and structures it in conversational format. 
During inference, \llminer first analyzes the importance of the sentence within a given context, and then employs this analysis in a chain-of-thought reasoning process to generate pertinent questions. Subsequently, it structures answers by synthesizing information from both the original text and the formulated questions. This procedure is repeated for each sentence in the document, thereby creating a multifaceted knowledge in instructional formats from the raw text. 
Unlike previous work largely depends on the LLM’s inherent ability to structure knowledge from raw text, \llminer offloads this knowledge structuring process to the training phase. We feed in multifaceted knowledge as training examples to enable LLMs to develop a more comprehensive understanding of domain-specific content.
These alleviate the problems including the model's failure to apply memorized text data to related queries \cite{zhu2023physics, sun2022recitation, wang2021can}, and its tendency to reiterate Wikipedia content without effectively responding to the query \cite{sun2023principle}.

For domain-specific chatbot training, we utilize \llminer generated data as two forms of augmentations: instructional formats for fine-tuning chatbots, and lists of answers that provide alternative perspectives to the original text. These augmentations, combined with the OpenAssistant Conversation Dataset (OASST), equip the chatbot with both conversational ability and domain-specific knowledge. In our findings, we observe that 1) The instructional data enhances the ability of extracting factual knowledge from raw text. 2) the list of answers as augmentation can enhance the understanding of document reflected in answering factual-based queries.

A key feature of \llminer is its minimal reliance on human intervention, requiring only a small set of seed instances generated by prompting GPT-4 to adapt an LLM to the knowledge mining process. This capability allows LLM to be easily adapted as a tool to help any AI agent to continuously evolve by integrating new domain knowledge.
Our contributions are highlighted in the following:
\begin{enumerate}
    \item We develop an LLM-based system for automatic knowledge mining, harnessing the model’s inherent natural language processing capabilities to extract and organize domain-specific knowledge.
    \item The data mined by \llminer offers a multi-perspective view of document content, facilitating a deeper comprehension of the original content.
    \item Our method demonstrates a pathway for LLM self-improvement using model-synthesized training data, enabling continuous adaptation and growth in line with the expanding landscape of domain knowledge.
\end{enumerate}

\section{Related Work}
\subsection{Question Generation}
Automatic question generation has been introduced to synthesize data for tasks such as question answering~\citep{narayan2023conditional, lewis2021paq, honovich2022true, honovich2021q, alberti2019synthetic}, retrieval tasks~\citep{dai2022promptagator} or dialogue systems~\citep{wang2023retrieval, dai2022dialog}. However, previous works use documents or specific spans within them as answer without re-writing. Furthermore, they require labeled QA data to train question generation. Models such PAQ requires training each model for each step in the generation pipeline, while our methodology harnesses a single LLM throughout the generation procedure.

\subsection{Instruction-tuning}
Recent endeavors to enhance LLMs have revolved around instruction tuning in QA~\cite{alpaca, sun2023principle, zhou2023lima} or conversation style~\cite{kopf2023openassistant,vicuna2023}. These efforts predominantly aim at aligning models with human preferences rather than bolstering knowledge acquisition, which requires large amount of knowledge-intensive quality QA data. Our paper presents such an effort in extracting high quality QA data from raw text as supervised data to encode knowledge into LLMs.

\subsection{Chain-of-thought Reasoning}
Chain-of-thought reasoning methods are proposed to improve the generation quality in various tasks~\cite{wei2022chain}. Techniques like ReciteLM~\cite{sun2022recitation} have been introduced, enabling LLMs to recite pertinent information during question answering (QA). However, these methodologies are tethered to the necessity of finely crafted in-context examples and exhibit shortcomings in incorporating external knowledge efficiently~\citep{sun2022recitation}.

\begin{figure*}[t!]
    \centering
    \includegraphics[width=\linewidth]{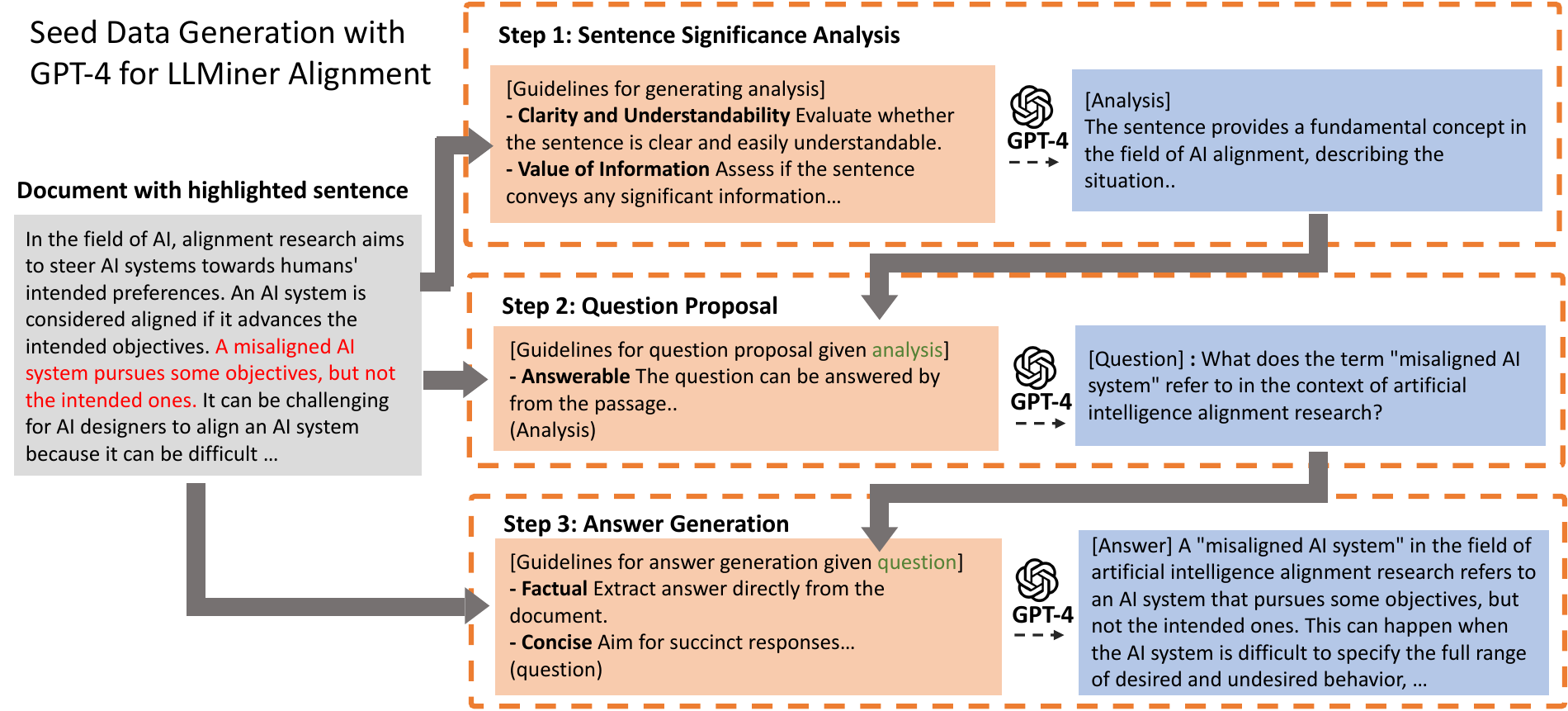}
    \caption{LLMiner Training Data Collection: Given a passage with a randomly selected sentence within it, we prompt GPT-4 with specific generation guidelines to analyze, question, and answer about the emphasized sentence in relation to the passage. The seed data is then applied for behavior cloning, fine-tuning a smaller LLM to mimic GPT-4's responses with the efficiency of shorter prompts.}
    \label{fig:llminer_illustration}
\end{figure*}


\section{Methodology}
In this section, we outline our approach for training a Large Language Model (LLM) into a specialized knowledge miner, which we refer to as \llminer. This model is designed to autonomously mine relevant Question-Answer (QA) pairs from specialized text corpora. These mined QA pairs are subsequently utilized to fine-tune the LLM, enhancing its proficiency in addressing domain-specific queries. The subsequent subsections will delve into three pivotal aspects of our methodology: 1) Preparing seed data using GPT-4 for training \llminer, 2) Fine-tuning a smaller LLM to align with the mining task, and 3) Employing \llminer to mine knowledge in instructional format.

\subsection{Preparing Seed Data with GPT-4}
Our aim in training \llminer is to enable it for versatile knowledge mining scenarios, capitalizing on the robust language understanding abilities inherent in pre-trained LLMs. To this end, we compile a diverse corpus that includes $300$ randomly sampled passages from Wikipedia, along with $50$ each from TREC-COVID, NFCorpus, ArguAna, FEVER, DBPedia, and SCIDOCS~\cite{thakur2021beir}, making for a total of \num instances.

As illustrated in \cref{fig:llminer_illustration}, for each passage, we start by randomly selecting a sentence to serve as the focal point. We then employ three prompts to generate an analysis, a question, and an answer, respectively. In each generation process, we adhere to a set of guidelines, along with a specific formatting approach as input for GPT-4~\cite{gpt4openai}. The full list of prompts is included in \cref{apd:gpt4_prompts}.

\paragraph{Significance Analysis}
We prompt GPT-4 to assess the significance of the selected sentence within the broader context of the entire passage. The guidelines for this analysis include factors such as clarity, importance, knowledge addition, and relevance to the overall content. 
The analysis is a chain-of-thought reasoning for organizing knowledge in the paragraph with the picked sentence as a focal point, serving as the initial step for question proposal. If the sentence is deemed unclear or uninformative, GPT-4 is required to output the specific guideline that has not been met, and we then bypass the subsequent steps for such sentences. These exceptions are used as training data to refine the model's filtering capabilities.

\paragraph{Question Proposal} 
If the sentence is determined to be significant, GPT-4 will be subsequently prompted to formulate questions aimed at encapsulating the insights obtained from the analysis. The guidelines specify that questions should be self-contained, answerable, and insightful. Given that the final deliverable from \llminer consists solely of QA pairs without accompanying passage context, it is crucial that the questions are formulated to be self-contained. To ensure this, our prompts direct the model to avoid using context-dependent phrases such as "from the document" or "in the report."

\paragraph{Answer Generation}
Lastly, GPT-4 proceeds to generate an answer to the previously formulated question. This phase engages the model in a reading comprehension task, adhering to guidelines that ensure the answer is factual, concise, and self-contained. By rephrasing or summarizing the information present in the original sentence, the model effectively reorganizes the information in the passage that aligns with the perspective of the generated question.

By following this three-step procedure, we gather a set of \num training instances from GPT-4 including the analysis, question and answer, which will be used to align an open-source LLM for the knowledge mining task. GPT-4 substitutes for human labeling efforts in this context.

\subsection{Training for Model Alignment}
Utilizing GPT-4 for the above knowledge mining comes with two primary drawbacks:
\begin{enumerate}
\item The computational overhead for large-scale inference is significant.
\item Each generative step necessitates detailed prompts that outline the guidelines for generating appropriate output.
\end{enumerate}
To address these issues, we employ a secondary, smaller LLM (LLaMa-7b) that we fine-tune to align with the GPT-4-generated seed data. 
The fine-tuning employs a simplified prompt (detailed in \cref{apd:llminer_prompts}), with a document and a highlighted sentence as input. The model outputs either an analysis or an analysis followed by a QA pair, depending on whether the sentence is important. 
This process effectively clones GPT-4's responses using more straightforward prompts.
The fine-tuning serves as a behavior cloning process to mimic the responses generated by GPT-4 using more concise and straightforward prompts.

\begin{figure*}[t!]
    \centering
    \includegraphics[width=\linewidth]{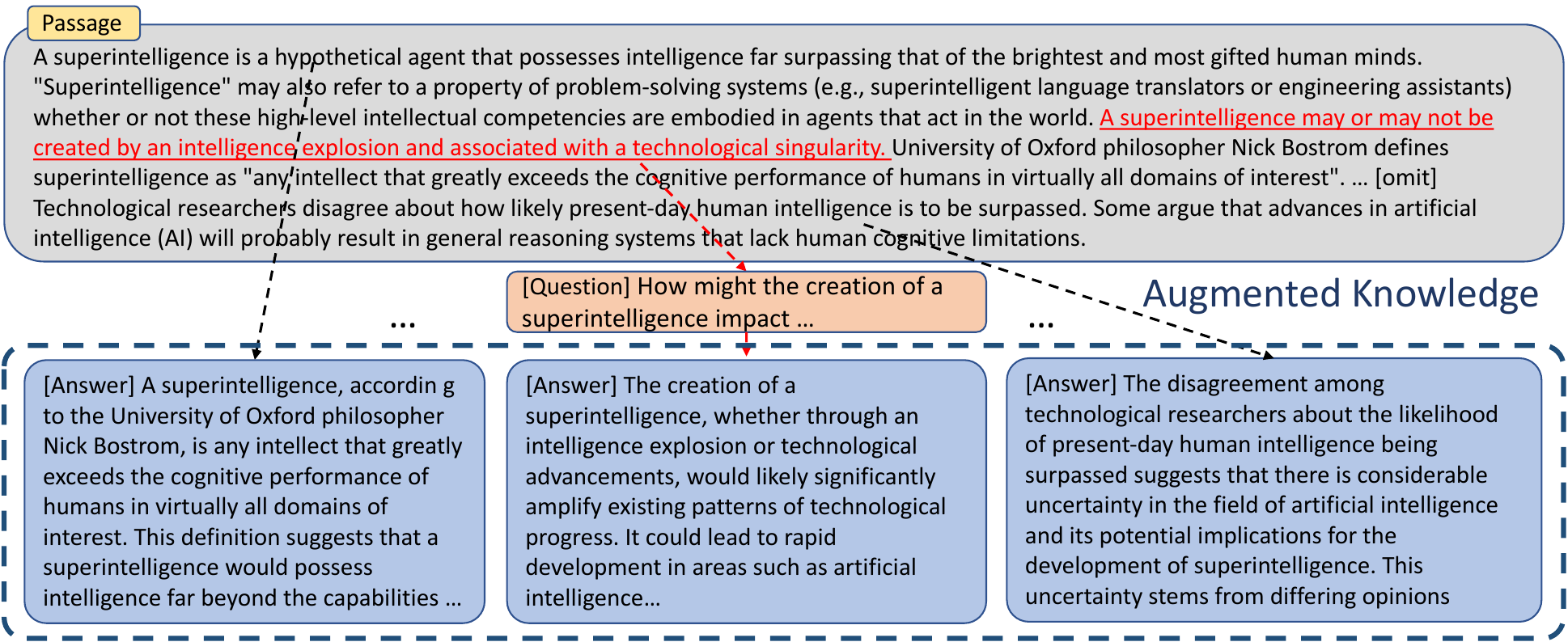}
    \caption{Knowledge Augmentation of the Original Passage: The collection of minded answers serves as augmented knowledge with each emphasizes specific facets determined by the highlighted sentence, enriching the understanding of the original passage.}
    \label{fig:augmented_knowledge}
\end{figure*}

\subsection{Inference with \llminer}
During the inference stage, we process each incoming passage by iterating through all its sentences, treating each one as a "highlighted sentence" for the model to focus on. We use the same prompt as in the training phase of \llminer for inference with each highlighted sentence within the passage. While we generate an intermediate analysis, we opt not to use it as knowledge representation. Instead, our primary goal is to mine knowledge in instructional format that can serve as training data for subsequent model fine-tuning.
Additionally, the answers from \llminer output also function as augmented knowledge in text form, adding richness and diversity to the original text corpus, as illustrated in Figure \cref{fig:augmented_knowledge}.

\section{Domain-Specific Text Collection}
This section outlines the methodologies for gathering domain-specific text and crafting test data. The objective is to facilitate the development of a specialized chatbot, tailored to user-defined topics such as "Artificial General Intelligence" or "Traditional Medicine in Southeast Asia." We delineate the processes for constructing a domain-specific corpus and formulating testing datasets.

\subsection{Construction of a Domain-Specific Corpus}
We employ two open-source tools, Wikipedia-API and GPT-4, to assemble a domain-specific corpus based on user-defined topics.

\paragraph{Initial Topic Expansion}
As a first step, we utilize GPT-4 to generate hypothetical Wikipedia titles related to the user-specified topic, as detailed in \cref{fig:corpus_prompt_a}. These titles, potentially non-existent on Wikipedia, act as initial search queries for the Wikipedia-API. Users may also contribute additional keywords to enhance these seed titles.

\paragraph{Wikipedia-Based Retrieval and Selection}
With the generated hypothetical titles and user-provided keywords in hand, we utilize the Wikipedia-API to retrieve actual Wikipedia titles.
Subsequently, GPT-4 is used to assess relevance, following the prompt specified in \cref{fig:corpus_prompt_b}. Articles corresponding to the relevant titles are incorporated into our domain-specific text corpus.

\subsection{Testing Dataset Creation}
This subsection elaborates on the approach to simulate real-world user interactions with the chatbot and the process of testing dataset creation, focusing on a scenario where the trainer customizes a chatbot for a specific topic, and users interact with it for inquiries within that domain.

\paragraph{Simulating User Queries}
To mimic user interactions, we engage annotators to propose questions that reflect potential user queries within the chosen topic. We aim to compile around 30 human-crafted questions per topic. These questions are designed to be factually based, targeting the chatbot's knowledge in the domain text. In addition, annotators are responsible for creating concise reference answers, typically one or two sentences long, for each question.

\paragraph{Independence from Training Corpus}
Annotators are allowed to utilize external resources like Google or ChatGPT to ensure the accuracy and relevance of their questions and answers. However, it's crucial to note that they do not have access to the actual text corpus or augmented data compiled by the chatbot trainer. This constraint ensures that the testing questions and responses are independent of the chatbot's training data, thereby allowing for an unbiased evaluation of the chatbot's performance in responding to unanticipated queries.

\section{Chatbot Training Methodology}
To train our specialized chatbot, we adopt a mixed-training approach that combines four distinct types of data:
\begin{enumerate}
\item \textbf{Raw Text:} This consists of the domain-specific text that we have gathered from Wikipedia, serving as a foundational source for domain expertise.
\item \textbf{Conversation Data:} We leverage the OpenAssistant Conversations Dataset (OASST) \cite{kopf2023openassistant} as a source of dialog interactions. To ensure quality, we only include entries with an overall score greater than 0.5, resulting in approximately 17k conversations.

\item \textbf{QA Pairs:} These are the question-answer pairs that we have mined using the methodologies described earlier. This data enriches the model's capability to respond to user queries in a focused manner.

\item \textbf{Augmented Knowledge:} In addition to using full QA pairs, we also include the list of mined answers as standalone elements to serve as augmented knowledge in text form. This approach aims to enhance the chatbot's reservoir of domain-specific information, as depicted in \cref{fig:augmented_knowledge}.
\end{enumerate}

By integrating these diverse data sources, we aim to create a robust and versatile chatbot capable of fielding domain-specific queries with a high degree of accuracy and contextual relevance. Since the mixing ratio of these different data types is not the primary focus of the paper,  we employ uniform sampling for the data types as our design choice. The training details are included in \cref{apd:training_details}.

\begin{figure*}[t!]
    \centering
    \includegraphics[width=\linewidth]{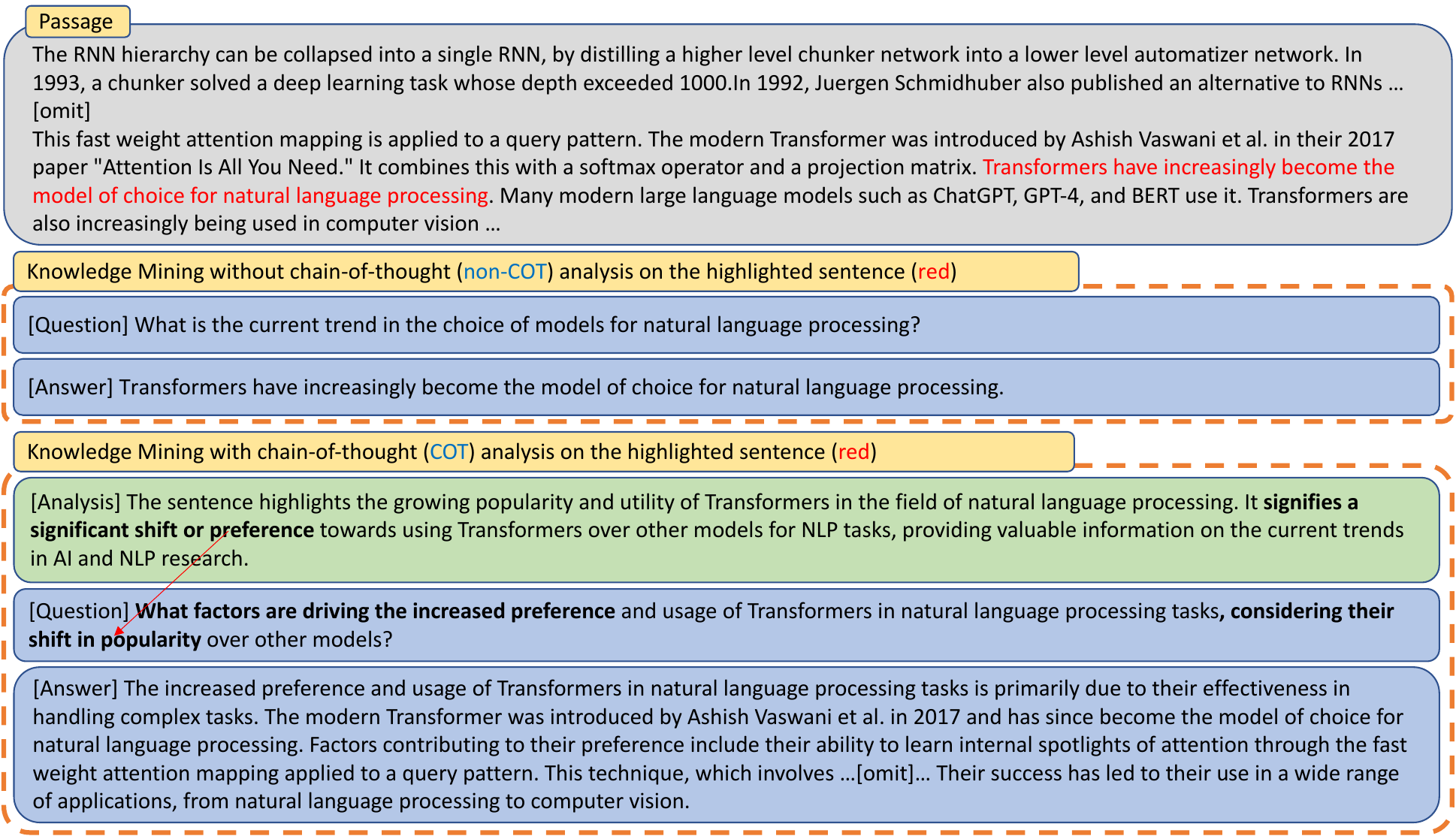}
    \caption{Example illustrating the utility of chain-of-thought (COT) analysis in extracting comprehensive knowledge from a passage. Without COT, the question can be directly answered from highlighted sentence, COT facilitates a deeper understanding by formulating context-rich questions referring to the analysis and synthesizing answers that requires an understanding the entire passage.}
    \label{fig:quality_comparison}
\end{figure*}

\section{Experiments and Evaluations}
This section delves into the empirical validation of \llminer's effectiveness by examining the quality of the generated Question-Answer (QA) pairs. Additionally, we assess the performance of a chatbot that utilizes \llminer, benchmarking it against several baseline models. These evaluations serve to substantiate the advantages of our method over prior approaches.

\subsection{Performance Evaluation of \llminer}
We assess \llminer's capabilities by comparing its output to a baseline knowledge miner that lacks chain-of-thought analysis. This baseline is modeled after existing methodologies as cited in \cite{dai2022promptagator, dai2022dialog, lewis2021paq}. For the sake of a balanced comparison, both \llminer and the baseline are trained using the same underlying model architecture and pipeline, albeit with distinct prompts, as detailed in Appendix \cref{apd:baseline_miner_prompts}.

\subsubsection{Qualitative Analysis}
An illustrative example of \llminer's performance is presented in Figure \cref{fig:llminer_illustration}. The figure underscores how \llminer emphasizes key pieces of information in the sentence while linking them to a broader context, thereby enriching the quality of the mined QA pairs.


\subsection{Chatbot Performance Evaluation}
We assess the performance of a chatbot trained on the LLaMa-7B model under various configurations:

\begin{itemize}
\item \textbf{Chatbot (Baseline)}: Trained exclusively on the OASST conversation dataset.

\item \textbf{Chatbot + Passage}: Serves as a major baseline and combines domain-specific passages with the OASST dataset.

\item \textbf{Chatbot + Passage + QA}: Integrates QA pairs generated by \llminer along with domain passages and the OASST dataset.

\item \textbf{Chatbot + Passage + QA + Augmentation}: Includes an augmented list of answers for knowledge enrichment, in addition to the components of the previous configuration.
\end{itemize}

We employ GPT-4 as a judge to rate the quality of the generated answers. The evaluations are conducted using a Likert scale ranging from 1 to 5, with prompts specified in Appendix \cref{apd:gpt_judge_prompts}. The results of this evaluation are presented in Table \cref{tab:chatbot_evaluation}.

Our findings indicate that the incorporation of QA data from \llminer can help learning domain knowledge from raw text corpus, which enhances the quality of chatbot to answer factual-based questions. Furthermore, mixing additional the training data with augmented knowledge (rewritten version of original passage by the list of answers) offers more perspectives of the original text, which improves the model's performance across three of the four topics assessed.

\begin{table*}[t]
\small
\setlength\tabcolsep{2pt}
    \centering
    \begin{tabular}{l | >{\centering\arraybackslash}p{2.5cm} >{\centering\arraybackslash}p{2.5cm} >{\centering\arraybackslash}p{2.5cm} >{\centering\arraybackslash}p{2.5cm}}
    \toprule
    & Artificial General Intelligence & Traditional Medicine in Southeast Asia & History of Steam Engine & Advances in Robotics Surgery \\
    \midrule
    Chatbot & 77.14 & 66.65 & 70.6 & 70.96 \\
    Chatbot + passage & 78.00 & 68.22 & 73.2 & 71.56 \\
    Chatbot + passage + QA & 80.67 & 68.92 & 79.8 & \bf 75.63 \\
    Chatbot + passage + QA + Aug & \bf 82.38  & \bf 70.47 & \bf 83.2 & 74.96\\
    \bottomrule
    \end{tabular}
    \caption{Performance comparison of different training setting for chatbot on the domain-specific texts, numbers are subjective to changes due to the on-going verification of QA testcase by human annotators.}
    \label{tab:chatbot_evaluation}
\end{table*}

\section{Conclusion}
We propose a methodology for transforming a Large Language Model (LLM) into a specialized knowledge miner, namely \llminer. Our approach leverages the power of GPT-4 to prepare seed data and fine-tunes a smaller LLM to perform the knowledge mining task efficiently. The mined Question-Answer (QA) pairs serve not only as enriched data for fine-tuning but also as a resource for domain-specific queries. Experimental evaluations demonstrated the effectiveness of our approach in both data mining quality and chatbot performance. Notably, the inclusion of QA pairs and additional text augmentations led to significant improvements in chatbot performance across multiple topics. Our findings underscore the potential of using LLMs as robust, domain-specific knowledge miners and shed light on the self-improvement of LLM training with automomous data mining.

\section{What will be released?}
We plan to release:
\begin{enumerate}
    \item The code for \llminer training, inference, and chatbot training. 
    \item The GPT-4 generated seed data for \llminer training and the LoRA weights of \llminer.
    \item The collection of domain corpus, mined knowledge data and the human annotated test set.
\end{enumerate}

%
%
%
\bibliography{custom}
\bibliographystyle{acl_natbib}

\newpage
\onecolumn
\newcommand{\smallstrut}{\vrule height 0.5\baselineskip depth 0pt width 0pt}
\appendix

\begin{figure*}[ht!]
    \centering
    \includegraphics[width=\linewidth]{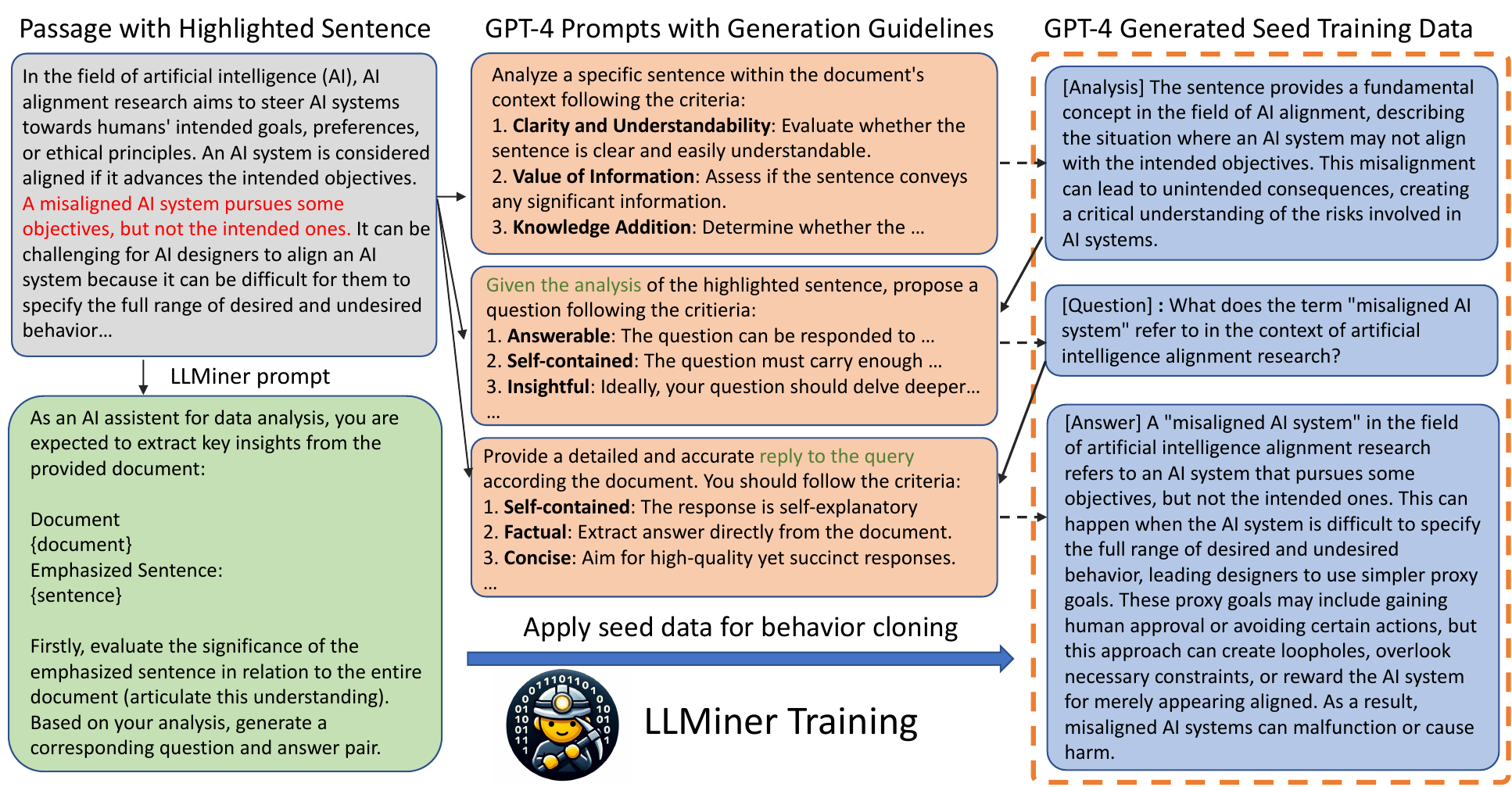}
    \caption{LLMiner Training Pipeline: The seed data is obtained by prompting GPT-4 prompts with specific generation guidelines to analyze, question, and answer about the emphasized sentence in a passage. The seed data is then applied for behavior cloning, fine-tuning a smaller LLM to mimic GPT-4's responses with the efficiency of shorter prompts.}
    \label{fig:llminer_pipeline}
\end{figure*}

\newpage
\section{GPT-4 Prompts for \llminer Seed Data Generation}
\label{apd:gpt4_prompts}
The overall pipeline of generating seed data is shown in \cref{fig:llminer_pipeline}. Specifically, the sentence analysis prompt is shown in \cref{fig:gpt4_analysis_prompt}, the question proposal prompt is shown in \cref{fig:gpt4_question_prompt} and the answer generation prompt is shown in \cref{fig:gpt4_answer_prompt}.

\begin{figure}[ht]
\centering
\small
\centering
\begin{minted}[fontsize=\footnotesize, frame=single,linenos=false,breaklines,breaksymbol=,escapeinside=||,bgcolor=Box1Color]{text}
Your role as an AI data miner requires you to extract pivotal details from a given document. To start, analyze a specific sentence within the document's context. Aim to evaluate the sentence's significance and provide a succinct summary of your findings. Use the following criteria to judge each sentence's importance:

Sentence Importance Evaluation:

1. Clarity and Understandability: Evaluate whether the sentence is clear and easily understandable.
2. Value of Information: Assess if the sentence conveys any significant information.
3. Knowledge Addition: Determine whether the sentence offers new insights or knowledge.
4. Relevance: Check whether the sentence delivers an essential message or key idea from the document.

Document Overview:

{document}

Sentence to Analyze:

"{sentence}"

Analysis Instructions:

Should a sentence meet the above criteria and is deemed important, your output should be "Yes." Then write a brief analysis of the sentence (two to three sentences). Concentrate on the sentence's informative content rather than restating its clarity or significance, i.e. "The sentence provides details in ..." is better that "The sentence is important as it provides ..." 

If the sentence fails to satisfy these criteria, your output should be "No," along with an explanation of why the sentence doesn't carry substantial importance.

Output:
\end{minted}
\caption{GPT-4 Prompt for Analysis Generation.}
\label{fig:gpt4_analysis_prompt}
\end{figure}

\begin{figure}[ht]
\centering
\small
\centering
\begin{minted}[fontsize=\footnotesize, frame=single,linenos=false,breaklines,breaksymbol=,escapeinside=||,bgcolor=Box1Color]{text}
As an AI data analyst, your mission involves mining crucial insights from a given document. To aid your analysis, a specific sentence from the document will be supplied, accompanied by an in-depth interpretation highlighting its importance. Using this interpretation, your task is to construct a question connected to the sentence.

Guidelines for framing your question:

1. Answerable: The question should be crafted such that it can be responded to using the context of the provided sentence.
2. Self-contained: Your question must carry enough context to be understood independently. If required, include an explanatory phrase for clarity. Avoid using terms like "from the document", "in the report".
3. Insightful: Ideally, your question should delve deeper than just surface-level details, making use of the analysis provided.

Document Brief:

{document}

Highlighted Sentence:

"{sentence}"

Analysis:

{analysis}

Given the document, highlighted sentence, and analysis, devise a question that can be comprehended without needing additional context - avoid phrasing like "in this context, ..." or "from the report ..."

Output:
\end{minted}
\caption{GPT-4 Prompt for Question Proposal.}
\label{fig:gpt4_question_prompt}
\end{figure}

\begin{figure}[ht]
\centering
\small
\centering
\begin{minted}[fontsize=\footnotesize, frame=single,linenos=false,breaklines,breaksymbol=,escapeinside=||,bgcolor=Box1Color]{text}
As an AI data analyst, your mission involves mining crucial insights from a given document. A text and an associated query will be provided. Your task is to articulate a detailed and accurate reply to the query, relying on the information embedded within the text. 

Response Crafting Instructions:

1. Self-contained: Ensure your response is self-explanatory and can be understood independently. While paraphrasing parts of the document for clarity is permitted, do not assume that the reader has prior knowledge of the document. Avoid using phrases such as "from the document", "in the report".
2. Factual: Extract your answer directly from the document. Avoid inventing facts if the query cannot be answered.
3. Concise: Aim for high-quality yet succinct responses. Ideally, a single paragraph will suffice.

Context:

{document}

Query:

{question}

Answer:
\end{minted}
\caption{GPT-4 Prompt for Answer Generation (Reading Comprehension).}
\label{fig:gpt4_answer_prompt}
\end{figure}

\newpage
\section{\llminer Prompts for Training and Inference}
\label{apd:llminer_prompts}
For \llminer alignment, we use a simpler prompt shown in \cref{fig:llminer_train_inf} with a passage and highlighted sentence as input and the GPT-4 generated seed data as output.
\begin{figure}[t]
\centering
\small
\centering
\begin{minted}[fontsize=\footnotesize, frame=single,linenos=false,breaklines,breaksymbol=,escapeinside=||,bgcolor=Box2Color]{text}
As an AI assistent for data analysis, you are expected to extract key insights from the provided document.

Document:

{document}

Emphasized Sentence:

{sentence}

Your assignment is to evaluate the significance of the emphasized sentence in relation to the entire document (articulate this understanding). Based on your analysis, generate a corresponding question and answer pair.
\end{minted}
\caption{\llminer prompt for training and inference.}
\label{fig:llminer_train_inf}
\end{figure}

\newpage
\section{Prompts for Building Domain Corpus}
\label{apd:corpus_prompts}
To build domain corpus, we employ GPT-4 to propose a list of hypothetical Wikipedia titles (prompt in \cref{fig:corpus_prompt_a}) as the first-round input. Then, we use the Wikipedia-API to search for real titles, and filter related titles with GPT-4 using prompt in \cref{fig:corpus_prompt_b}.
\begin{figure}[t]
\centering
\small
\centering
\begin{minted}[fontsize=\footnotesize, frame=single,linenos=false,breaklines,breaksymbol=,escapeinside=||,bgcolor=Box1Color]{text}
Can you give a list of hypothetical Wikipedia titles to search for the topic "{topic}"?

Output format:
A python array [List of Wikipedia titles]

Output:
\end{minted}
\caption{Prompt for proposing hypothetical Wikipedia titles with GPT-4, the titles will be used as input to Wikipedia-API to search for relevant documents.}
\label{fig:corpus_prompt_a}
\end{figure}

\begin{figure}[t]
\centering
\small
\centering
\begin{minted}[fontsize=\footnotesize, frame=single,linenos=false,breaklines,breaksymbol=,escapeinside=||,bgcolor=Box1Color]{text}
Determine if the list of searched document titles are related to the query of "{topic}".

List of searched document titles:
{titles}

Output format:
Related titles: [List of related titles]

Output:
\end{minted}
\caption{Prompts of employing GPT-4 to filter for relevant titles from the search results of Wikipedia-API.}
\label{fig:corpus_prompt_b}
\end{figure}

\newpage
\section{Training Hyperparameters}
\label{apd:training_details}
\begin{figure}[t]
\centering
\small
\centering
\begin{minted}[fontsize=\footnotesize, frame=single,linenos=false,breaklines,breaksymbol=,escapeinside=||,bgcolor=Box3Color]{bash}
# single GPU training with batch size 4
python3 -m lmlib.trainer.train_lora \
    --model_name_or_path decapoda-research/llama-7b-hf \
    --data_path $data_path --conv_template conv_llminer \
    --cache_dir $CACHE_DIR \
    --output_dir $output_dir \
    --lora_target_modules 'q_proj,k_proj,v_proj,o_proj,gate_proj,up_proj,down_proj' \
    --lora_r 64 --lora_dropout 0.05 \
    --group_by_length \
    --bf16 True \
    --num_train_epochs 4 \
    --per_device_train_batch_size 4 --gradient_accumulation_steps 1 \
    --evaluation_strategy "no" \
    --logging_steps 25 \
    --save_strategy "epoch"\
    --save_total_limit 5 \
    --learning_rate 5e-4 \
    --warmup_ratio 0.03 \
    --lr_scheduler_type "cosine" \
    --model_max_length 2048 \
    --gradient_checkpointing \
    --q_lora True
\end{minted}
\caption{Hyperparameters for alignment of \llminer.}
\label{fig:llminer_hyperparam}
\end{figure}

\begin{figure}[t]
\centering
\small
\centering
\begin{minted}[fontsize=\footnotesize, frame=single,linenos=false,breaklines,breaksymbol=,escapeinside=||,bgcolor=Box3Color]{bash}
# actual batch size is 8 x 4 x 2 = 64
torchrun --nproc_per_node=8 --master_port=$port lmlib/trainer/train_mem.py \
--model_name_or_path $model_name \
    --data_path $data_path --conv_template vicuna_v1_1 \
    --cache_dir $CACHE_DIR \
    --output_dir $model_output_dir \
    --bf16 True \
    --num_train_epochs 3 \
    --per_device_train_batch_size 4 --gradient_accumulation_steps 2 \
    --evaluation_strategy "no" \
    --logging_steps 25 \
    --save_strategy "epoch"\
    --save_total_limit 2 \
    --learning_rate 2e-5 --lr_end 2e-6 --weight_decay 0.01 --warmup_ratio 0.03 --lr_scheduler_name "linear_warmup_cosine"  \
    --model_max_length 2048 \
    --fsdp "full_shard auto_wrap" \
    --fsdp_transformer_layer_cls_to_wrap 'LlamaDecoderLayer' \
    --gradient_checkpointing True \
    --fsdp_config "fsdp_config/config.json" \
    --optim adamw_torch 
\end{minted}
\caption{Hyperparameters for chatbot training.}
\label{fig:chatbot_hyperparam}
\end{figure}
For \llminer training, we use Qlora~\cite{dettmers2023qlora} to optimize the base LLaMa-7b model for $4$ epochs with the  hyperparameters shown in \cref{fig:llminer_hyperparam}

For training Chatbot, we use full-model training to finetune all parameters with the hyperparameters shown in \cref{fig:chatbot_hyperparam}

\newpage
\section{Baseline Miner Prompts for GPT-4 and Knowledge Minder Training}
\label{apd:baseline_miner_prompts}
For comparison of LLMiner with and without a chain-of-thought step (similar setting to previous works~\cite{lewis2021paq,dai2022promptagator,dai2022dialog}), we repeat the process of data collection only using a different set of prompts. The prompt for seed data generation with GPT-4 is shown in \cref{fig:gpt4_baseline_prompt} and the prompt for baseline \llminer training and inference is shown in \cref{fig:llminer_baseline_prompts}.

\begin{figure}[t]
\centering
\small
\centering
\begin{minted}[fontsize=\footnotesize, frame=single,linenos=false,breaklines,breaksymbol=,escapeinside=||,bgcolor=Box1Color]{text}

As an AI data analyst, your task is to extract essential information from a provided document and transform a highlighted sentence into a question and answer (QA) pair.

QA generation guidelines:

1. Meaningful: Evaluate whether the sentence conveys a clear message. If ambiguous or unclear, provide a rationale for its exclusion.
2. Answerable: Construct the question such that the answer can be directly inferred from the given context.
3. Standalone: The crafted question should be comprehensive on its own. Incorporate any necessary context or explanatory phrases to ensure the question is self-sufficient.

Document Context:

{document}

Highlighted Sentence:

"{sentence}"

If the sentence is coherent and meaningful, format your output as:

Question: <proposed question>
Answer: <answer derived from the document>

Should the sentence lack clarity or significance, elucidate the reason for its omission:

Skip: <reason for skipping>

Output:
\end{minted}
\caption{GPT-4 Prompt for Question-Answer proposal without a chain-of-thought step.}
\label{fig:gpt4_baseline_prompt}
\end{figure}

\begin{figure}[t]
\centering
\small
\centering
\begin{minted}[fontsize=\footnotesize, frame=single,linenos=false,breaklines,breaksymbol=,escapeinside=||,bgcolor=Box2Color]{text}
As an AI assistent for data analysis, you are expected to extract key information from the provided document.

Document:

{document}

Emphasized Sentence:

{sentence}

From the document and the emphasized sentence, generate a pertinent question and answer pair.
\end{minted}
\caption{A baseline knowledge miner prompt without chain-of-thought reasoning step, used as comparison for \llminer.}
\label{fig:llminer_baseline_prompts}
\end{figure}

\newpage
\section{GPT-4 Judge Prompts}
\label{apd:gpt_judge_prompts}
We evaluate the quality of LLM output with GPT-4 judge with prompt shown in \cref{fig:llm_judge_prompt}.
\begin{figure}[t]
\centering
\small
\centering
\begin{minted}[fontsize=\footnotesize, frame=single,linenos=false,breaklines,breaksymbol=,escapeinside=||,bgcolor=Box1Color]{text}
Please act as an impartial and objective judge and evaluate the quality of the response provided by a chatbot to the user's question. Your evaluation should be mainly based on whether the response is correct, and whether the response contains any hallucinations.

To evaluate the LLM responses, first, begin your evaluation by providing a short explanation. Second, after providing your explanation, you must rate the response by choosing from the following options:

1 - Completely Incorrect, irrelevant, with hallucination
2 - Mostly Incorrect, with hallucination
3 - Somewhat Incorrect / Partially Correct, with hallucination
4 - Mostly Correct but with some hallucination
5 - Correct, no hallucination

Question:
{question}

Reference Answer:
{answer}

LLM Response to Evaluate:
{llm_response}

Output format:
Explanation: <your explanation>
Rating: <your rating>
\end{minted}
\caption{Prompt for using GPT-4 as LLM judge.}
\label{fig:llm_judge_prompt}
\end{figure}

\newpage
\section{\llminer Examples}

{ \footnotesize \begin{tabularx}{\linewidth}{r X} \toprule 
Passage & 
\textcolor{red}{Most surveyed AI researchers expect machines to eventually be able to rival humans in intelligence, though there is little consensus on when this will likely happen.} At the 2006 AI@50 conference, 18\% of attendees reported expecting machines to be able "to simulate learning and every other aspect of human intelligence" by 2056; 41\% of attendees expected this to happen sometime after 2056; and 41\% expected machines to never reach that milestone. In a survey of the 100 most cited authors in AI (as of May 2013, according to Microsoft academic search), the median year by which respondents expected machines "that can carry out most human professions at least as well as a typical human" (assuming no global catastrophe occurs) with 10\% confidence is 2024 (mean 2034, st. dev. 33 years), with 50\% confidence is 2050 (mean 2072, st. dev. 110 years), and with 90\% confidence is 2070 (mean 2168, st. dev. 342 years). These estimates exclude the 1.2\% of respondents who said no year would ever reach 10\% confidence, the 4.1\% who said 'never' for 50\% confidence, and the 16.5\% who said 'never' for 90\% confidence. Respondents assigned a median 50\% probability to the possibility that machine superintelligence will be invented within 30 years of the invention of approximately human-level machine intelligence. In a 2022 survey, the median year by which respondents expected "High-level machine intelligence" with 50\% confidence is 2061. The survey defined the achievement of high-level machine intelligence as when unaided machines can accomplish every task better and more cheaply than human workers. In 2023, OpenAI leaders published recommendations for the governance of superintelligence, which they believe may happen in less than 10 years.
\\
\toprule \texttt{Analysis} $\to$ &
The sentence introduces the main topic of the document, which is the expectation of AI researchers regarding the future of machine intelligence. It highlights that while a majority of these researchers believe machines will eventually equal human intelligence, there is a lack of consensus on the timeline for this event. This sets the stage for the following discussion on various predictions and estimates related to AI advancement. 
\\
\noalign{\vskip 2mm}  \texttt{Question} $\to$ &
What does the consensus among AI researchers indicate about the timeline for machines achieving equal intelligence to humans, and how does this relate to the lack of a definitive answer?
\\
\noalign{\vskip 2mm} \texttt{Answer} $\to$ &
Most surveyed AI researchers anticipate machines achieving equal intelligence to humans within the next few decades to a couple of decades. However, there is a wide range of expectations regarding the exact timeline. A 2006 AI@50 conference survey revealed that 18\% of attendees expected machines to simulate human intelligence by 2056, 41\% expected it to happen sometime after 2056, and 41\% believed it would never happen. In a 2022 survey, respondents assigned a median 50\% probability to the possibility that machine superintelligence will be invented within 30 years of the invention of approximately human-level machine intelligence. OpenAI leaders in 2023 published recommendations for the governance of superintelligence, stating that they believe it may happen in less than 10 years. The lack of a definitive answer might be attributed to the varying levels of confidence in these predictions and the inherent uncertainties in predicting future AI advancements.
 \\
\toprule \texttt{Non-COT QA} $\to$ &
Question:
What do most surveyed AI researchers expect about the ability of machines to rival humans in intelligence?

Answer:
Most surveyed AI researchers expect machines to eventually be able to rival humans in intelligence, though there is little consensus on when this will likely happen.
 \\
 \bottomrule \end{tabularx} }
\end{document}